\documentclass[10pt,conference]{IEEEtran}
\IEEEoverridecommandlockouts
\usepackage{cite}
\usepackage{amsmath,amssymb,amsfonts}
\usepackage{algorithmic}
\usepackage{graphicx}
\usepackage{textcomp}
\usepackage[dvipsnames]{xcolor}
\usepackage{listings}
\usepackage[utf8]{inputenc}
\usepackage{color}
\usepackage{tabularray}
\usepackage{graphicx}
\usepackage{colortbl}
\usepackage{threeparttable}
\usepackage{multirow}
\usepackage{fancybox}
\usepackage{setspace}
\usepackage{caption}
\usepackage{subcaption}
\usepackage{algorithm}
\usepackage{algorithmic}
\usepackage[most]{tcolorbox}
\usepackage{xcolor}
\usepackage{adjustbox}
\usepackage{nicematrix}
\usepackage{multirow}
\usepackage{threeparttable}
\usepackage{hyperref}
\usepackage{balance}
\usepackage{enumitem}
\usepackage{booktabs}
\usepackage{verbatim}
\usepackage{collab}
\usepackage{tikz}
\usepackage{forest}
\usetikzlibrary{trees, arrows.meta, positioning, shadows}

\def\BibTeX{{\rm B\kern-.05em{\sc i\kern-.025em b}\kern-.08em
    T\kern-.1667em\lower.7ex\hbox{E}\kern-.125emX}}

\definecolor{grey}{rgb}{0.9, 0.9, 0.9}

\collabAuthor{yt}{red}{Yintong Huo}
\collabAuthor{yc}{blue}{Yichen Li}
\collabAuthor{rf}{orange}{Ruofan Lu}

\begin{document}

\title{Exploring Autonomous Agents: A Closer Look at Why They Fail When Completing Tasks}
%
\author{
  \IEEEauthorblockN{
    Ruofan Lu\IEEEauthorrefmark{2},
    Yichen Li\IEEEauthorrefmark{2},
    Yintong Huo\IEEEauthorrefmark{3}\IEEEauthorrefmark{1}
  }


  \IEEEauthorblockA{\IEEEauthorrefmark{2}The Chinese University of Hong Kong, Hong Kong.
    Email: \{rflu, ycli21\}@cse.cuhk.edu.hk}

  \IEEEauthorblockA{\IEEEauthorrefmark{3}Singapore Management University, Singapore.
    Email: ythuo@smu.edu.sg}
}

\maketitle

\begin{abstract}
Autonomous agent systems powered by Large Language Models (LLMs) have demonstrated promising capabilities in automating complex tasks. However, current evaluations largely rely on success rates without systematically analyzing the interactions, communication mechanisms, and failure causes within these systems. To bridge this gap, we present a benchmark of 34 representative programmable tasks designed to rigorously assess autonomous agents. Using this benchmark, we evaluate three popular open-source agent frameworks combined with two LLM backbones, observing a task completion rate of approximately 50\%. Through in-depth failure analysis, we develop a three-tier taxonomy of failure causes aligned with task phases, highlighting planning errors, task execution issues, and incorrect response generation. Based on these insights, we propose actionable improvements to enhance agent planning and self-diagnosis capabilities. Our failure taxonomy, together with mitigation advice, provides an empirical foundation for developing more robust and effective autonomous agent systems in the future.
\let\thefootnote\relax\footnotetext{$^{*}$ Yintong Huo is the corresponding author.}
\end{abstract}

\begin{IEEEkeywords}
LLM agents, autonomous agents, failure analysis.
\end{IEEEkeywords}

\section{Introduction}

The advancement of LLMs has enabled a new trend in task automation through autonomous agents~\cite{qiao2023taskweaver,wu2023autogen,zhang2024codeagent,wang2024executable}. These agents are designed to work together to interpret human commands, autonomously produce and execute code, and return the answer directly to the user. This synergistic workflow is capable of resolving more complicated problems, and provides an ``end-to-end'' answer without user involvement in the technical coding processes.

\begin{figure}[t]
    \centering
    \includegraphics[width=\columnwidth]{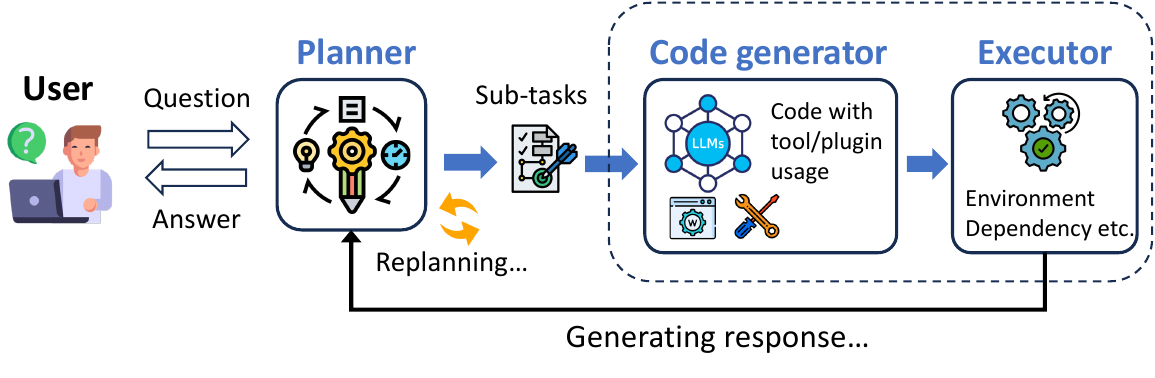}
    \vspace{-0.25in}
    \caption{The basic framework of an autonomous agent system.}
    \label{fig:framework}
\vspace{-0.2in}
\end{figure}

Current agent systems are implemented as a collaborative team of specialized LLMs abstracted into three core components (Figure~\ref{fig:framework}): (1) Planner, who decomposes complex user requests into a sequential plan of tasks, (2) Code generator, which converts each sub-task into executable and functional code with the use of various tools or plugins, and (3) Executor, which runs the code and integrates with development environments. The executor collects outputs and errors, forming a feedback loop to the planner for refining or returning an answer. Some studies refer to the combined code generation and execution process as the code interpreter~\cite{zhou2023solving, zhang2024cibench, low2023data, zhang2025webpilot}.

Despite the promising capabilities of agent systems, their performance remains merely understood beyond a basic success rate metric~\cite{guo2024redcode}.  There is an absence of systematic analysis in exploring the intricate communication among agents, their information passing mechanism, and the root causes of failures. For instance, whether a failure in web crawling task stems from incorrect planning or code generation.
To advance these systems in the long run, it is essential to identify the fundamental bottlenecks by conducting a thorough investigation into failure origins.

To fill this gap, we built a benchmark containing 34 representative programmable tasks to evaluate current autonomous agents. Using this benchmark, we assessed three widely-used open-source agent frameworks paired with two LLM backbones. 
Our experiments show that approximately 50\% of tasks are successfully completed by current agent collaborations, with failure causes including improper task planning, generation of nonfunctional code, and inadequate refinement strategies across iterations. 
Based on this analysis, we propose a three-level taxonomy of failure causes aligned with different task phases. Additionally, we offer several actionable recommendations aimed at enhancing planning capabilities and self-diagnosis mechanisms to advance autonomous agents.

We summarize the contribution of this paper as follows:
\begin{itemize}[leftmargin=*]
    \item \noindent \textit{Benchmark}: We build a benchmark with programmable tasks to evaluate the capabilities of current autonomous agents.
    \item \noindent \textit{Failure analysis}: Our evaluation of three popular agent frameworks reveals an approximately 50\% task completion rate, categorizing failures into a three-level taxonomy based on task phases.
    \item \noindent \textit{Actionable advice}: We provide suggestions to improve planning and self-diagnosis mechanisms, aimed at improving future autonomous agent systems.
\end{itemize}

\begin{table*}[t]
\centering
\footnotesize
\caption{A description of the evaluation of agent frameworks in terms of their design goals and collaborative strategies.}
\vspace{-0.1in}
\label{tab:framework-characteristics}
\begin{tabular}{l||p{6cm}p{8cm}}
\toprule
    \textbf{Framework} & \textbf{Design Goals} & \textbf{Collaboration Strategy} \\
\midrule
    TaskWeaver & Translating user requests into executable code for task automation. & A stateful and linear workflow of plan generation, coding for each step, and an interpreter executes it.\\
\midrule
    MetaGPT & Generate projects simulating a software development company. & Encoding standard operating procedures into prompt sequences, following an assembly line to pass information to each other to finish complicated tasks. \\
\midrule
    AutoGen & A flexible framework for agents to solve tasks via conversation. & Adapted from flexible agent conversations, agents chat with each other, forming a dynamic and interactive collaboration to complete tasks.\\
\bottomrule
\end{tabular}
\vspace{-0.15in}
\end{table*}

\section{Autonomous Agents in Software Engineering}
Recent research has increasingly applied LLM-based agents to software engineering from two directions~\cite{liu2024large}: developing agents to handle specific SE tasks~\cite{he2025llm, xia2024agentless} (e.g., debugging), and improving agent frameworks by enhancing role definitions and collaboration mechanisms~\cite{dong2024self, qiao2023taskweaver}. 
Within these directions, agent-based approaches have demonstrated promising results across key SE domains measured by task success rates, such as requirement engineering~\cite{jin2024mare,wei2024requirements}, code generation~\cite{liu2024marscode,ding2024cycle,jiang2024self} and testing~\cite{mundler2024swt,liu2024make,chen2024chatunitest}. However, most prior work treats agent systems as monolithic entities and lacks in-depth analysis of their internal processes. 
In contrast, our study emphasizes a detailed analysis from the perspective of the agent framework itself, investigating the contributions and interactions of individual agents, exposing current limitations, and guiding the design of more effective, collaborative agent systems in the future.


\section{Experiments}

\subsection{Benchmark construction}
We selected three types of common coding tasks in daily life for our benchmark, with their sources as follows:
\begin{itemize}[leftmargin=*]
    \item \textbf{Web Crawling}: We search for the keyword “Web Crawling” on GitHub and Stack Overflow and construct tasks from the returned repositories and posts.
    \item \textbf{Data Analysis}: We incorporate a number of tasks from DABench~\cite{DAbench}, an end-to-end data analysis benchmark that requires agents to interact with an executable code environment to solve problems.
    \item \textbf{File Operations}: We curate tasks based on several Stack Overflow posts focusing on fundamental file operations using Python and Bash.
\end{itemize}
During task selection, we follow the criteria below to ensure benchmark quality.
First, tasks have to be executable, with evaluation based on the outcomes of running the code rather than the code alone, differentiating our approach from typical code generation benchmarks.
Second, tasks need to accommodate automated evaluation, leading us to exclude tasks like front-end interface generation that are challenging to assess programmatically.
Finally, tasks are verified to be at least partially solvable by an agent, enabling meaningful exploration of design challenges; overly difficult or completely unsolvable tasks were left out. We carefully construct the benchmark consisting of 34 tasks, with human-verified ground-truth labels for automatic evaluation.

\textit{Metrics.} Following prior work~\cite{bogin2024super, yao2024tau}, we measure success rate as the evaluation metric. A task is deemed successful only if its output exactly matches the ground-truth answer.

\subsection{Studied subjects}
Since our research focuses on the agent frameworks rather than the capabilities of the underlying LLMs, we select three popular open-source mainstream frameworks to examine their agent interconnectivity and collaboration mechanisms: TaskWeaver~\cite{qiao2023taskweaver}, MetaGPT~\cite{hong2023metagpt}, and AutoGen~\cite{wu2023autogen}. Table~\ref{tab:framework-characteristics} presents their brief descriptions with design goals and collaboration strategies.
For the LLM backbones embedded within each agent, we select GPT-4o~\cite{gpt4o} and GPT-4o mini~\cite{gpt4omini} for evaluation.


\subsection{Implementation}
After the task selection is completed, we designed a general prompt template to standardize requests across different categories of tasks, which contains the Task Description, Instruction, Constraints, and Environment Information.

We implemented the benchmark as a toolbox enabling automated execution and evaluation. The agent frameworks were deployed on a Linux server, each running within its respective containers and sandboxes, using Python 3.10.14. We used MetaGPT 0.8.1, AutoGen 0.2.36, and TaskWeaver (commit hash number cf76c3b70b29ef64185fd3c9af0510c9e2fcc51e).
For the agent backbones, we employed two models: GPT-4o (gpt-4o-2024-08-06) and GPT-4o mini (gpt-4o-mini-2024-07-18). Results from task executions underwent post-processing and information extraction to facilitate efficient automated evaluation, while full logs were documented for later analysis. 

\section{Result analysis}
This section presents evaluation results and summarizes the failure causes into a taxonomy according to experiments.

\begin{figure}[t]
    \centering
    \vspace{-0.15in}
    \includegraphics[width=0.9\columnwidth]{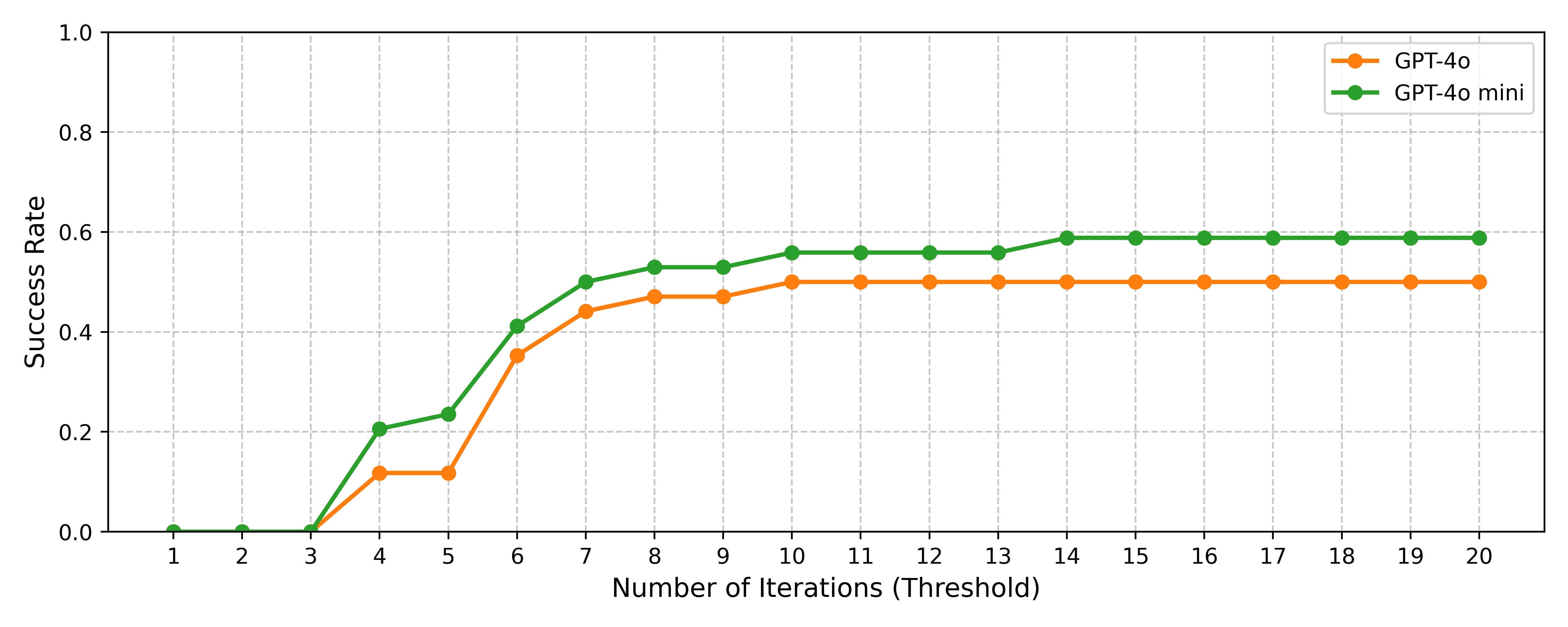}
    \vspace{-0.05in}
    \caption{The success rate concerning max iteration numbers.}
    \vspace{-0.1in}
    \label{fig:iteration}
\end{figure}

\begin{table}[t]
\centering
\footnotesize
\caption{Benchmark success rate (\textbf{GPT-4o}).}
\resizebox{\columnwidth}{!}{%
\begin{tabular}{l||cccc}
\toprule
\textbf{Agent} & \textbf{Web Crawling} & \textbf{Data Analysis} & \textbf{File Operations} & \textbf{All} \\
\midrule
TaskWeaver & 16.67 & 66.67 & 75.00 & 50.00 \\
MetaGPT    & 33.33 & 55.56 & 50.00 & 47.06 \\
AutoGen    & 16.67 & 50.00 & 50.00 & 38.24 \\
\bottomrule
\end{tabular}%
}
    \vspace{-0.1in}
\label{tab:agent-performance-4o}
\end{table}

\subsection{Quantitative analysis}

\begin{figure*}[ht]
    \centering
    \vspace{-0.1in}
    \includegraphics[width=0.9\textwidth]{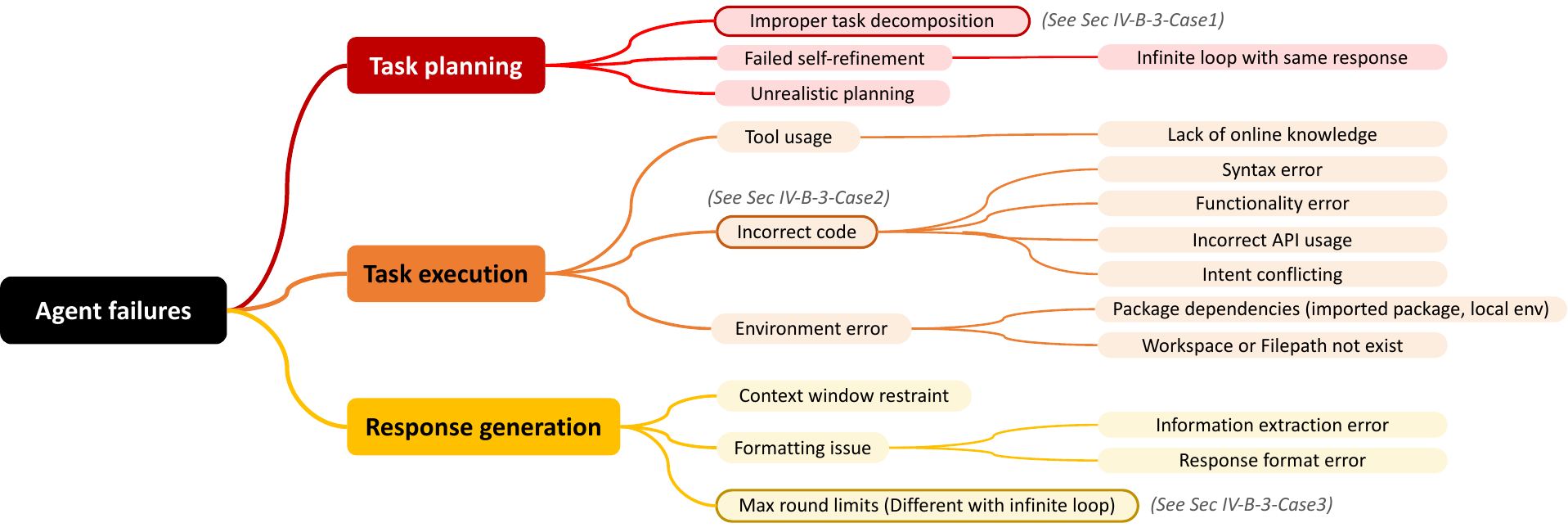}
    \caption{The agent failure taxonomy, where the most frequent failure is highlighted and illustrated in Sec.~\ref{sec:common-failure-analysis}.}
    \label{fig:taxonomy}
    \vspace{-0.1in}
\end{figure*}


Table~\ref{tab:agent-performance-4o} and \ref{tab:agent-performance-4omini} presents the evaluation results on GPT-4o and GPT-4o-mini with discussions as follows.

\textbf{Agent performance decreases on reasoning-intensive tasks.}
Agent performance varies by task. Using GPT-4o, agents perform well in Data analysis and File operations, with TaskWeaver scoring 67\% and 75\%, respectively. Web crawling is more challenging, with scores as low as 17\%, due to its reasoning-intensive nature, requiring code generators to infer element paths from user intent and HTML data. Compared to unstructured web tasks, simpler, structured tasks like data analysis benefit more from autonomous agents.

\begin{table}[t]
\centering
\footnotesize
\caption{Benchmark success rate (\textbf{GPT-4o mini}).}
\vspace{-0.05in}
\resizebox{\columnwidth}{!}{%
\begin{tabular}{l||cccc}
\toprule
\textbf{Agent} & \textbf{Web Crawling} & \textbf{Data Analysis} & \textbf{File Operations} & \textbf{All} \\
\midrule
TaskWeaver & 50.00 & 55.56 & 100.00 & 58.82 \\
MetaGPT    & 25.00 & 66.67 & 50.00 & 50.00 \\
AutoGen    & 41.67 & 44.44 & 100.00 & 50.00 \\
\bottomrule
\end{tabular}%
}
\label{tab:agent-performance-4omini}
\vspace{-0.1in}
\end{table}

\textbf{A cross-agent comparison reveals distinct specializations.}
With GPT-4o, TaskWeaver leads in structured tasks such as Data analysis (67\%) and File operations (75\%), while MetaGPT excels in Web crawling (33\%). Both TaskWeaver and AutoGen achieve perfect scores in File Operations, highlighting their architectures' strong compatibility with the lightweight model for executing precise, procedural tasks.

\textbf{While stronger models have higher reasoning capabilities, they might run into the overthinking issue and compromise results.}
In our evaluation, both TaskWeaver and MetaGPT show strong results with GPT-4o, scoring 50.00\% and 47.06\% overall. Surprisingly, the smaller GPT-4o-mini outperforms, especially in web crawling tasks, indicating that simpler models can remain highly competitive. Analysis of execution logs reveals that GPT-4o’s failures stem from a conflict between its task-planning processes (such as requesting additional confirmations) and built-in safety constraints (like denying web scraping), causing it to produce valid plans but then halt execution. Such ``overthinking'' ultimately results in task failure. The superior performance of GPT-4o-mini is consistent with prior research findings~\cite{schnabel2025multi}.



\textbf{More iterations improve success, but with diminishing gains after a certain threshold.}
Figure~\ref{fig:iteration} shows the success rate over the threshold of iterations in TaskWeaver. The success rate is zero for the first two iterations, indicating that a minimum number of attempts is necessary to solve the tasks. Between iterations 3 and 10, there is a rapid improvement in the success rate for both models, with the most significant gains occurring in this phase. After 10 iterations, increasing the maximum number of iterations yields only marginal gains.

\subsection{Failure study}
\subsubsection{Manual investigation}
The experiments were conducted across 34 tasks, 2 LLM backbones, and 3 agent frameworks, totaling 204 runs, with 104 task failures recorded along with detailed experimental logs. We recruit three authors, each with a minimum of two years of programming experience, to review the agent execution logs. These logs include comprehensive information such as prompt construction for each agent, individual agent outputs, and execution results for every iteration.

Our investigation began with the first-level failure taxonomy, where all annotators agreed to categorize failures according to the roles of key phases: task planning, task execution, and response generation. Next, each annotator independently reviewed the failure logs to summarize second-level failure reasons. Finally, they collaboratively discussed their categorizations and reached consensus on the final taxonomy.

\subsubsection{Failure taxonomy}
Figure~\ref{fig:taxonomy} presents the failure taxonomy, which encompasses 19 distinct causes across three tiers.

\textbf{Task planning.} A planner is responsible for breaking down user instructions into a sequence of executable sub-tasks for the code generator. This role is critical since the planner’s output directly guides subsequent agents and largely determines the success of the overall framework. We identified three common issues in planning: (1) improper task decomposition that generates steps that are logically incorrect or unsuitable for the assigned task; (2) failed self-refinement involves the model is unable to learn from its past errors, causing it to repeat the same failed sub-tasks in an infinite loop; (3) unrealistic planning refers to producing a sequence of plausible steps but exceeds the practical capabilities of downstream agents, making the sub-tasks impossible to execute.

\textbf{Task execution.} Task execution is the phase where the agent attempts to carry out the planned sub-tasks, involving the failures from the code generator and executor. Existing agent frameworks encounter three main failures: (1) the generator agent fails to exploit external tools (e.g., available functions), often due to a lack of online or tool-use knowledge. (2) the generator agent produces flawed code with syntax errors, functionality errors (executable but deviating from the intended output), incorrect API usage with wrong parameters, or showing conflicts to its original goal, and (3) executions also fail with improper environmental setup, such as missing dependency package and accessing a file that does not exist.

\begin{figure}[t]
    \centering
    \includegraphics[width=0.9\columnwidth]{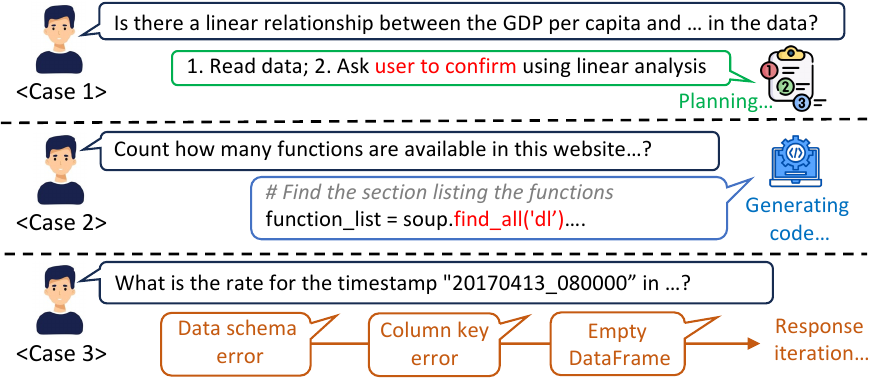}
    \caption{Three most common failure types.}
    \vspace{-0.2in}
    \label{fig:cases}
\end{figure}

\textbf{Response generation.}
Response generation is the final stage where the agent produces output for the user or the planner to use in subsequent iterations. Failures at this stage relate to how results are perceived and presented, even after the code has been executed. Three main failures causes are: (1) context window restraint: the agent loses parts of the conversation, leading to responses that are disconnected from previous interactions (e.g., an overly large HTML file in a web crawling task), (2) formatting issue: the agent’s output contains irrelevant information or does not comply with the required format (e.g., returning a sentence when a number is expected), (3) maximum rounds exceeded: The agent reaches a preset limit on the number of interaction turns without successfully completing the task, despite attempting various plans.

\subsubsection{Common failure analysis}\label{sec:common-failure-analysis}
For the most common failure in phases (i.e., planning, execution, and response generation), we present one case for each in Figure~\ref{fig:cases} and analyse as follows. 

\textbf{Case 1:}
The user asks the agent to verify a linear relationship between data. However, instead of proceeding directly to generate the necessary code for the analysis, the planner adds a redundant step: asking the user for confirmation to use linear analysis, though such usage has been specified in task description. This unnecessary clarification introduces a bottleneck, halting the process until user feedback is provided.
Such redundant planning not only delays the task but also degrades the user experience by creating unnecessary interaction.

\textbf{Case 2:}
When tasked with counting the number of functions on a website, the agent generates code that operates on an incorrect assumption. The code \texttt{soup.find\_all('dl')} presumes that all \texttt{$<$dl$>$}  HTML tags on the page are used exclusively for listing functions. However, on complex webpages like technical documentation, these tags are often used for a variety of purposes, including navigation, definitions, or other structural elements. This flawed assumption leads to an incorrect count and demonstrates a failure to understand the contextual use of HTML structure, resulting in faulty code.

\textbf{Case 3:}
The agent fails when trying to find a specific data point. It first gets a \texttt{KeyError} due to an additional space in a column name. The agent then switches to an alternative strategy of retrieving the entire row, which also fails in \texttt{Empty DataFrame}. Such an error implies that the agent faces challenges in self-correcting based on the output of its previous checks. It therefore leads to a loop of failures that ultimately exceeds the maximum attempts and causes the task to fail.

\section{Actions on mitigating agent failures}

The failures analyzed highlight critical weaknesses in agent systems, particularly in planning and error correction. To address these, we propose two key strategies as follows.


\textbf{Promoting planning ability with learning-from-feedback.}
Planner is the first and fundamental component of an autonomous agent, decomposing complex tasks into executable steps. We therefore advocate for a ``learning-from-feedback'' design, where agents learn to re-plan from their previous operational environment feedback. Recent work shows that agents can dynamically adjust plans based on tool feedback, deciding whether to refine or restart~\cite{yao2023react, hao2023reasoning} the pre-defined plan, avoiding rigid and illogical steps. Such a feedback-aware mechanism also shows promise in software engineering applications like program repair~\cite{xia2024automated} and code generation~\cite{peng2025perfcodegen, zhang2023self}. This allows the agent to adapt new strategies when faced with unexpected outcomes.


\textbf{Developing early-stop and navigation mechanism.}
Failures like infinite loops and hitting round limits highlight the agent's inability to recover from repeated mistakes. 
To this end, future agent systems can develop a meta-controller that navigates to a certain agent upon root cause analysis, either re-planning to correct a strategic error or invoking a specialized tool to fix a local execution fault. 
Proper navigation can efficiently fix the problem, reducing task attempts and improving reliability.
Moreover, if the system detects repetitive, unresolved errors, the mechanism should trigger an ``early stop'', halting the process before it hits the maximum round limit, thereby saving resources.


\section{Conclusion and Future Work}
This study investigates autonomous agent systems, focusing on their collaboration mechanisms and the reasons behind their failures in completing end-to-end tasks. We evaluate three popular agent frameworks using a newly developed benchmark, analyze their outcomes, and classify the causes of failure. Additionally, we propose two practical design strategies for agent frameworks to address common failure modes. 
In the future, we aim to enrich the benchmark and implement these strategies.
Our benchmark data and evaluation framework can be found at \url{https://github.com/lurf21/Agent_Evaluation_Framework}.

\section{Acknowledgment}
 We appreciate all the anonymous reviewers for their valuable and practical comments. The work  was supported by the
Singapore Ministry of Education (MOE) Academic Research Fund (AcRF) Tier 1 grant.

\clearpage
\balance
\bibliographystyle{IEEEtran}
\bibliography{ref}

\end{document}